\documentclass{article}
\usepackage{ijcai26}

\usepackage{times}
\usepackage{soul}
\usepackage{url}
\usepackage[hidelinks]{hyperref}
\usepackage[utf8]{inputenc}
\usepackage[small]{caption}
\usepackage{graphicx}
\usepackage{amsmath}
\usepackage{amsthm}
\usepackage{booktabs}
\usepackage{algorithm}
\usepackage{algorithmic}
\usepackage[switch]{lineno}

\usepackage{hyperref}
\usepackage{url}
\usepackage{graphicx}
\usepackage{booktabs}
\usepackage{multirow}
\usepackage{xspace}
\usepackage{tcolorbox}
\usepackage{listings}
\usepackage{bm}
\usepackage{amsfonts}

\def\mytool{\textsc{PaLRS}\xspace}
\def\mytooln{Preference alignment of Large Language Models via Residual Steering\xspace}

\title{Toward Preference-aligned Large Language Models\\via Residual-based Model Steering}

\author{
Lucio La Cava
\and
Andrea Tagarelli\\
\affiliations
DIMES Dept., University of Calabria, Italy\\
\emails
\{lucio.lacava, tagarelli\}@dimes.unical.it
}

\begin{document}

\maketitle

\begin{abstract}
Preference alignment is a critical step in making Large Language Models (LLMs) useful and aligned with (human) preferences. Existing approaches such as Reinforcement Learning from Human Feedback or Direct Preference Optimization typically require curated data and expensive optimization   over billions of parameters, and eventually lead   to persistent task-specific models. 
In this work, we introduce \mytooln (\mytool), a training-free method that exploits preference signals encoded in the residual streams of LLMs. From as few as one hundred preference pairs, \mytool extracts lightweight, plug-and-play steering vectors that can be applied at inference time to push models toward preferred behaviors. 
We evaluate \mytool on various small-to-medium-scale  open-source LLMs, showing that \mytool-aligned models achieve consistent gains on mathematical reasoning and code generation benchmarks while preserving baseline general-purpose performance. 
Moreover, when compared to models aligned with DPO and SimPO, they perform better with great time-savings. 
Our findings highlight that \mytool offers  an effective, much more efficient and flexible alternative to standard preference optimization pipelines, offering a training-free, plug-and-play mechanism for alignment with minimal data.
\end{abstract}

\vspace{-4mm}
\section{Introduction}
Large Language Models (LLMs) have rapidly advanced the state-of-the-art performance across various domains, including dialogue, programming, and mathematical tasks~\cite{LLM2025surveycapabilities}.
While most capabilities in such systems are due to rich and wide pretraining~\cite{chen-etal-2024-parallel,lmd3colm,shaib-etal-2024-detection,wang2025iclr-pretraining}, a key determinant in their usability is how close their outputs align with \textit{human preferences}~\cite{llmalignsurvey2023,llmalignsurvey}.
Indeed, preference alignment has emerged in recent years as a focal stage in the LLM deployment pipeline: approaches such as reinforcement learning from human feedback~\cite{humanfeedbacknips22,humanfeedback}, direct preference optimization (DPO)~\cite{dpo}, or simple preference optimization (SimPO)~\cite{simpo} have become   standard practices for eliciting better capabilities from LLMs.

Despite their tangible effects,   preference-optimization alignment methods remain costly and inflexible. 
First, current approaches rely on large volumes of curated preference datasets~\cite{openassistant}, thus being highly \textit{annotation intensive}. %
Second, despite parameter-efficient methods such as LoRA adapters, aligning a model remains \textit{computationally intensive} because it requires repeated forward and backward passes through large models, often consuming several GPU-hours~\cite{stiennon2020learning}. 
Third, alignment is typically considered \textit{persistent}: once an LLM has been fine-tuned toward a particular preference setting, adapting it to a different set of preferences generally requires starting again from the base model to produce new checkpoints. Maintaining multiple preference-specific checkpoints can quickly become resource-intensive. 
These challenges underscore the need to scale alignment methods across three dimensions: efficiency (reducing computational cost), effectiveness  (maximizing alignment quality), and flexibility (enabling rapid adaptation to new preference specifications).

Recently, an emerging line of research has unveiled that \textit{residual stream activations} of LLMs encode contextually rich and linearizable features that can be used to manipulate the model behavior surgically, yet without altering their weights or requiring any additional  training~\cite{representation-engineering,Zhang24patching}.
Residual-based interventions have been shown effective to mitigate refusal behaviors~\cite{Arditi24refusal,Wang25surgical}, erase concepts~\cite{Belrose23leace}, induce desired persona-like behaviors~\cite{chen2025persona}, or improve factfulness~\cite{li2024truthful}.
These results suggest a promising direction, with residuals functioning as “control knobs,” providing a relatively inexpensive yet effective mechanism for steering model behavior at inference time. However, prior work has focused on optimizing narrow behaviors (e.g., refusal mitigation), leaving the broader challenge of using residual interventions to align models with a range of preferences underexplored.

\begin{figure}[t!]
    \centering
    \setlength{\tabcolsep}{2pt}
    \begin{tabular}{cc}
        \includegraphics[width=0.48\linewidth]{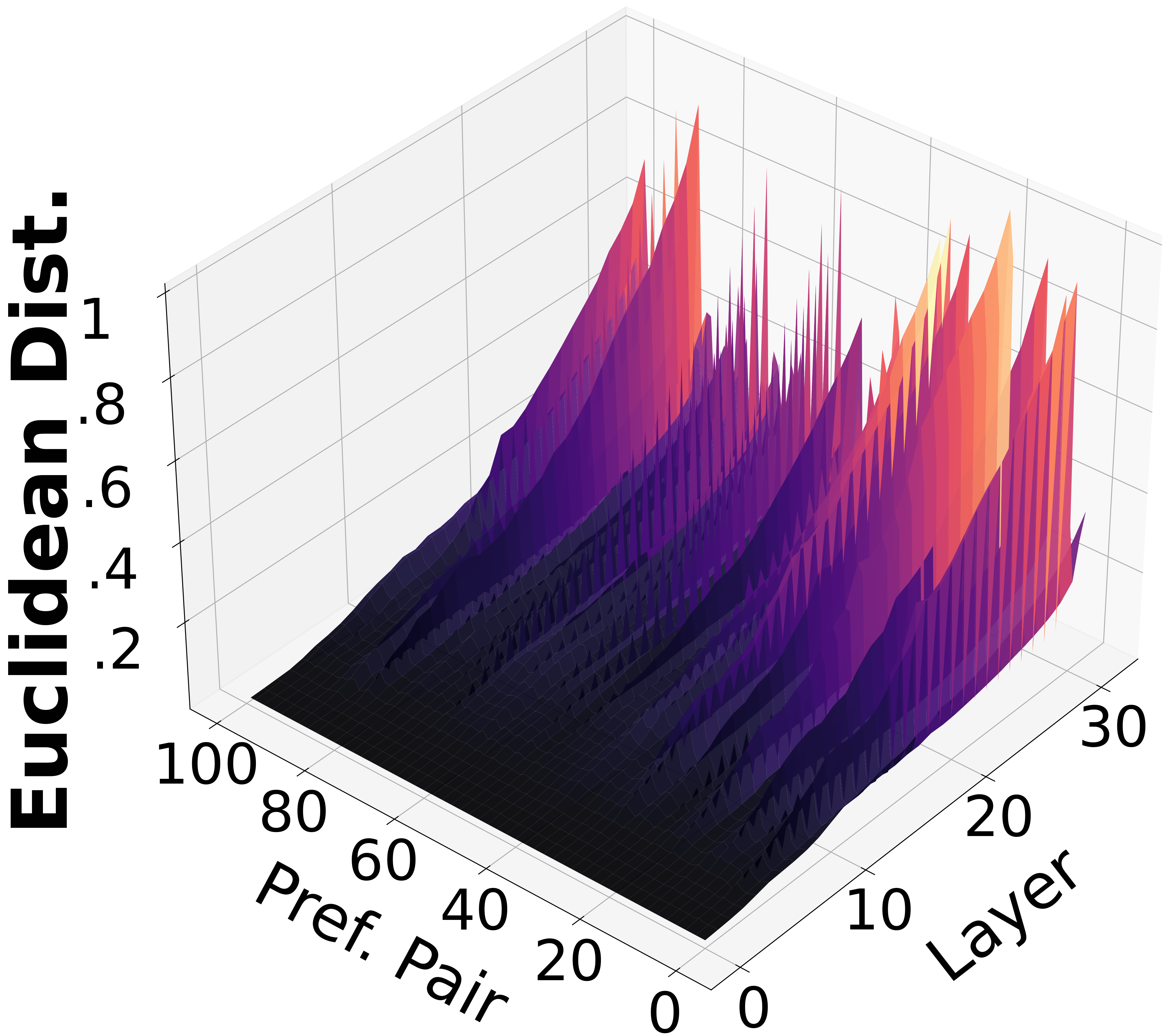} &
        \includegraphics[width=0.48\linewidth]{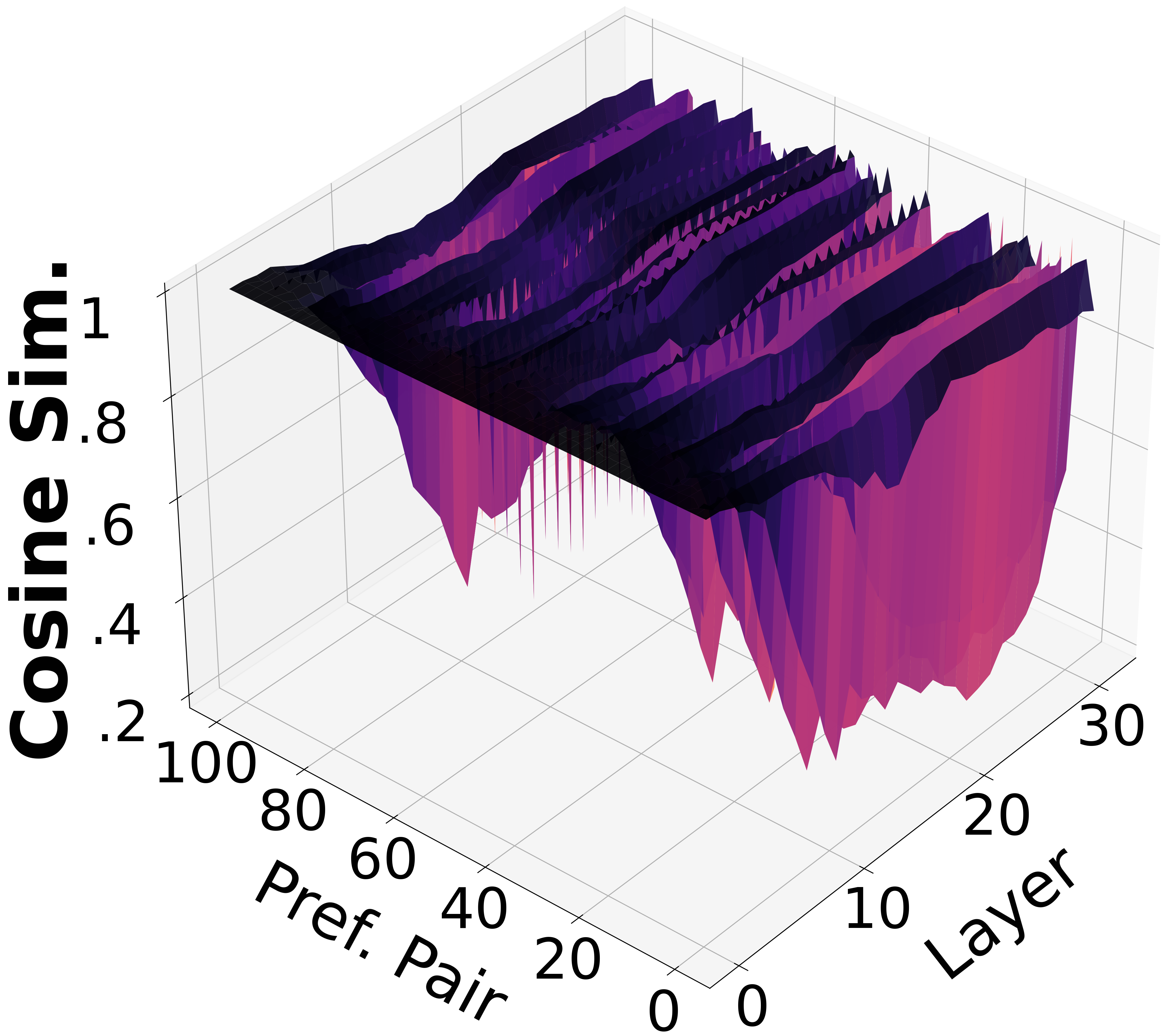}
    \end{tabular}
    \caption{Normalized Euclidean distance  and cosine similarity between residual stream activations (from Llama 3.1 8B Instruct) of 100 randomly-sampled chosen-rejected response pairs to math questions, for a   fixed token position (cf. Sect.~\ref{subsec:extraction}) and by varying the model layer ids---higher ids are closer to the output layer. Lighter colors denote higher Euclidean distance and lower cosine similarity.
    }
    \label{fig:motivating}
\end{figure}

\paragraph{Our Hypothesis.} 
Consider the task of improving a model's mathematical reasoning or coding capabilities. Standard practices, based on preference optimization alignment, would require curating thousands of preference pairs, optimizing the model's weights, and deploying task-specific checkpoints, resulting in a resource-intensive  and inflexible process.

In this work, we adopt a different approach, as we hypothesize that the difference between the residual stream activations of chosen and rejected responses---to questions grounded in a  particular domain, e.g., math or coding---can be meaningful. If so, we could distill these differences into steering directions that can be used to induce \textit{preferred behaviors} in a model via lightweight inference-time interventions. 

To support our  hypothesis, consider  Fig.~\ref{fig:motivating}, which shows  Euclidean distance and cosine similarity between  residual stream activations  of preference pairs (i.e., chosen-rejected responses) to math questions. 
It is worth noticing that   the Euclidean distances and cosine similarities can  substantially vary across the pairs. In particular, there is a significant portion of pairs for which the Euclidean distance  resp. cosine similarity increases resp. decreases with mid-high layer ids. 
This implies the possibility of defining  a vector, or steering direction,  that moves the activation from the rejected toward the chosen response: if the difference were tiny, a steering vector would have little effect; by contrast, large distances suggest the difference is substantial enough to guide model behavior adjustment. In addition, the evidence of cases with low cosine similarity indicates that the differences between those pairs' activations are mostly consistent in direction, thus allowing   a generalizable aggregated steering vector to be distilled, rather than needing a separate vector for each example.

\paragraph{\mytooln \ (\mytool).}
Building on the above remarks, we introduce a novel approach to preference alignment of LLMs through steering with residual stream activations, dubbed \mytool. 
Instead of updating model weights via gradient optimization, \mytool computes preference directions based on  differences in residual activations extracted from  a small set of preference pairs (on the order of 100). These directions are then applied at inference time, enabling lightweight, plug-and-play     preferred-behavior steering.

To the best of our knowledge, this is the first study to leverage residual stream directions for preference alignment. Our contributions are threefold:

$\bullet$  We bring the model steering framework based on residual stream activations to the setting  of preference   alignment of LLMs, showing that  residual stream activations encode preference information in a linearly accessible form. 

$\bullet$ We introduce a simple difference-in-means approach for estimating candidate preference directions from a small set of chosen–rejected response pairs, without requiring any LLM post-training. We further propose a principled criterion for  selecting the steering direction that best aligns an LLM’s behavior with target preferences.
 
$\bullet$ We demonstrate the effectiveness  of  \mytool through different small-to-medium scale open-source LLMs, with testing on widely used benchmarks.  Particularly, as a concrete case study, we  show that steering directions derived from preference-alignment datasets conceived for \textit{math reasoning} (\textsf{GSM8K}) and \textit{code generation} (\textsf{HumanEval}),  under certain conditions of the steering intensity,  lead an LLM to improve performance on corresponding benchmarks, without degrading results on other-domain benchmarks.  
Additionally, we highlight that \mytool-aligned models take substantial advantage w.r.t.  DPO- and SimPO-aligned models on both \textsf{GSM8K} and \textsf{HumanEval}, achieving superior effectiveness while requiring much less computational overhead.

\vspace{1.5mm}
\noindent
\textbf{Impact of our research. }
A key advantage of our method lies in the efficiency afforded by its training-free strategy. \mytool requires only a single forward pass   over few (e.g., 100) data samples to create the steering vector, which accounts for its substantial computational speedups. In contrast, standard preference-optimization alignment methods rely on full fine-tuning procedures, where data is a critical bottleneck and (tens of) thousands of paired samples are typically required.
As a result, \mytool is markedly data-efficient, making it particularly attractive for low-resource scenarios, including sensitive domains such as medical applications. 

The flexibility of \mytool is another notable strength. The approach is not restricted to specific domains such as mathematical reasoning or code generation. Because \mytool operates via task-specific steering vectors rather than permanent model modifications, multiple vectors (e.g., for math, coding, or other tasks) can coexist and be applied as needed. This plug-and-play generality highlights the potential of \mytool as a broadly applicable framework for preference alignment.

\section{Related Work}

\paragraph{Model Steering via Feature Directions.}
Recent studies suggest that linear directions in the activation space of LLMs capture richer and more generalizable features than individual neurons~\cite{li-etal-2021-implicit,superposition}. 
This shift toward subspace-level aspects has motivated research that exploits linear representations to probe model internals and provide better interpretation and steering of their behaviors~\cite{hernandez-andreas-2021-low,nanda-etal-2023-emergent,Park24linear,bayat2025steering}. 
A key challenge is to reliably identify such feature directions. In this regard, unsupervised approaches based on Sparse Auto-Encoders (SAE) have been used to uncover latent, interpretable features~\cite{Huben24sae,lan2024quantifying,sae-survey}. A complementary trend involves exploiting contrastive pairs of texts differing across a specific axis to extract directions that isolate targeted behaviors~\cite{Burns23unsupervised}.
Once extracted, these features can be used to intervene on model activations, particularly in the residual stream, where editing is known to effectively steer the models' behavior~\cite{representation-engineering,Zhang24patching,wang-etal-2025-beyond-prompt}. 
Such interventions have been shown promising in shifting sentiment and detoxification~\cite{turner2023steering}, enhancing truthfulness~\cite{li2024truthful}, erasing concepts~\cite{Belrose23leace}, targeting refusal behaviors~\cite{Arditi24refusal,Wang25surgical}, and controlling character traits~\cite{chen2025persona}.

\paragraph{Preference Optimization in LLMs.}
A large body of work has focused on aligning LLMs with human preferences, aiming to render such tools more usable~\cite{llmalignsurvey2023,llmalignsurvey}. 
This has been largely driven by Reinforcement Learning from Human Feedback (RLHF) approaches~\cite{humanfeedbacknips22,humanfeedback}, and more efficient methods like Direct Preference Optimization (DPO)~\cite{dpo}. 
Nonetheless, despite recent efforts to further improve efficiency ~\cite{hong-etal-2024-orpo,simpo}, the promising and more efficient alternative of steering models toward preferences via residual streams remained almost   unexplored.

In a related effort,  recent  studies~\cite{liu-etal-2024-aligning,rimsky-etal-2024-steering,Cao24bipo} explore preference alignment by identifying disparities in activation patterns elicited by preferred vs.  dispreferred stimuli. While these represent an important step toward bridging preference data with internal model representations, their approaches differ fundamentally from ours. 
 \cite{liu-etal-2024-aligning}, resp.~\cite{Cao24bipo} requires training with contrastive stimuli to extract the relevant signals, resp. optimizing the bi-directional objective, whereas our method is entirely \textit{training-free}. Also, \cite{liu-etal-2024-aligning} introduces a low-rank adaptation module to perform steering, in contrast to our \textit{inference-time intervention}. In addition, \cite{rimsky-etal-2024-steering} mostly focuses on contrastive prompt pairs with multiple-choice questions. 
Overall, the above  design choices incur limitations  in practice, with    more limited steering effects,  less generalizable preferences, and  less scalable and lightweight improvements compared to \mytool.

\section{Methodology}

\subsection{Preliminaries}
\label{sec:preliminaries}

\paragraph{Notation.}
Throughout this paper, we will use capital letters to denote data objects, lowercase letters to denote scalars, and bold lowercase letters to denote vectors. 

We are given a collection of text triplets   $\langle Q, A^{(+)}, A^{(-)}\rangle$ where $Q$ is a question, and $A$s are two possible (human-provided) answers to $Q$.  
Based upon this collection, we define the dataset  
$\mathcal{D} = \{ \langle Q, A^{(+)}, A^{(-)}\rangle \, | \, A^{(+)} \succ_{Q} A^{(-)} \}$,  where $A^{(+)} \succ_{Q} A^{(-)}$ denotes that, for question $Q$, $A^{(+)}$ is preferred over $A^{(-)}$. 
Let also $\mathbf{t}^{(+)}$,  $\mathbf{t}^{(-)}$, and  $\mathbf{t}^{(Q)}$ denote the input sequences of tokens from  an answer $A^{(+)}$,   $A^{(-)}$  and   question $Q$, respectively.

\paragraph{Residual stream activations. }
Following previous studies~\cite{Zhang24patching}, we resort to the concept of \textit{residual stream activation}, specifically in the context of     the Transformer decoder architecture, with $L$ layers and hidden size $d$.   
Given a   sequence $\mathbf{t}$, the state of knowledge a model has about a token in position $i$ at the start of layer $\ell$ (i.e., before layer $\ell$ processes it)    can   be expressed by the token's residual stream activation, given the input tokens up to position $i$ and all contributions computed from the  layers preceding $\ell$.
We will denote it as a real-valued $d$-dimensional vector  $\mathbf{x}_{i,\ell}(\mathbf{t})$, or simply with $\mathbf{x}_{i,\ell}$ if $\mathbf{t}$ is clear from the context.

In other words, the token's residual stream activation is the accumulated hidden representation of the token as it flows through the model, thus encoding all contextual information that the model has built up so far---which includes  the initial token and positional embeddings, plus the  contributions from the multi-head causal self-attention and feed-forward MLP components (sublayers) of all previous layers. This also means that the residual stream is linearly interpretable, since sublayer outputs are added to the residual stream, and is tied to the model's prediction distribution, since at the final layer the model projects the last residual stream through the embedding-to-token matrix to get the next-token logits.

\paragraph{Model steering. }
The residual stream activation can be used for model steering because it is the central state vector that encodes all of the model’s knowledge about a token at a given point. 

One effective way to  reliably alter the model's outputs and behavior is \textit{activation addition}, i.e.,   adding an identified direction $\mathbf{r} \in \mathbb{R}^d$ in the  residual space that corresponds to some feature, e.g., tokens that are more  representative of  the chosen responses than the rejected ones. Therefore, shifting the residual stream along that direction will change the model's predictions accordingly.  

An opposite approach is \textit{directional ablation}, which consists in subtracting from the residual stream  its orthogonal projection onto the direction $\mathbf{r}$. However, as noted in  ~\cite{Arditi24refusal}, the  two approaches have a different impact as the directional ablation applies to all layers and token positions; by contrast, the activation addition involves only a desired layer (and applies across all token positions). 
While ablation has been proven effective in refusal-removal setups~\cite{Arditi24refusal,Wang25surgical}, it is not well-suited to our setting: indeed, chosen and rejected responses typically differ only by a few key tokens, thus ablating their direction is very likely to disrupt  the model's behavior.

\subsection{Extracting Preference Directions}
\label{subsec:extraction}
To extract the candidate \textit{preference directions} from the model's residual stream  activations, we resort to the \textit{difference-in-means} approach, which is proven to be worst-case optimal~\cite{belrose2023diffinmeans}, i.e., no linear method can achieve lower worst-case error in recovering a linearly encoded concept direction, and has  shown to be effective in previous work~\cite{Arditi24refusal,hollinsworth-etal-2024-language,Wang25surgical}. 

Given the dataset $\mathcal{D} = \{ \langle Q, A^{(+)}, A^{(-)}\rangle \, | \, A^{(+)} \succ_{Q} A^{(-)} \}$  of triplets of questions and chosen-rejected responses, 
 we first compute two quantities for any choice of token position $i$ and layer $\ell$, which correspond to the  average of the residual stream activations produced by the model when it receives in input  the chosen, resp. rejected, responses:
\begin{equation}
\begin{aligned}
    \bm{\mu}_{i,\ell}^{(+)} = \dfrac{1}{|\mathcal{D}|} \sum_{A^{(+)} \in \mathcal{D}} \mathbf{x}_{i,\ell}(\mathbf{t}^{(+)}) \\
    \bm{\mu}_{i,\ell}^{(-)} = \dfrac{1}{|\mathcal{D}|} \sum_{A^{(-)} \in \mathcal{D}} \mathbf{x}_{i,\ell}(\mathbf{t}^{(-)}).
\end{aligned}
\end{equation}
 
We define the \textit{candidate preference direction} for a given token position $i$ and layer $\ell$ as the difference between the two means as follows:  
\begin{equation}
    \mathbf{r}_{i,\ell} = \bm{\mu}_{i,\ell}^{(+)} - \bm{\mu}_{i,\ell}^{(-)}.
\end{equation}

It is worth noticing that, to prevent diluting the residual stream signals, we focus on the chosen and rejected responses while discarding   the instruction $Q$ from the computation of directions.  Indeed, we are interested in discerning the signals that differentiate the chosen tokens from the rejected ones, regardless of a particular prompt.

\begin{table}[t!]
    \centering
    \scalebox{0.8}{
    \begin{tabular}{c|p{0.45\textwidth}}
    \toprule
    \textbf{Model} & \multicolumn{1}{c}{\textbf{Template}} \\
    \midrule
        Llama & {\raggedright\ttfamily{<|begin\_of\_text|>\\<|start\_header\_id|>user\\<|end\_header\_id|> {\textit{instruction}}\\{\color{blue}<|eot\_id|><|start\_header\_id|>\\assistant<|end\_header\_id|>}}}\\
        \midrule
        Mistral & {\raggedright\ttfamily{<s>[INST] {\textit{instruction}} {\color{blue}[/INST]</s>}}}\\
        \midrule
        OLMo & {\raggedright\ttfamily<|endoftext|><|user|>\\{\textit{instruction}} {\color{blue}<|assistant|>}}\\
    \bottomrule
    \end{tabular}
    }
    \caption{Single-turn chat templates for the model families considered in this study. Blue-colored text indicates \textit{post-instruction tokens}. Note that a single special string may be decomposed into multiple token IDs by the model’s tokenizer, and for the sake of readability non-visible tokens (e.g., newlines) have been omitted.}
    \label{tab:templates}
\end{table}

\paragraph{Position and layer selection strategies.}  
Following~\cite{Arditi24refusal}, we  consider only token positions corresponding to \textit{post-instruction tokens}, i.e., the template tokens following the instruction, ensuring the model processed the given text and starts producing its output (Table~\ref{tab:templates}). 

Regarding the selection of layers, we emphasize the importance of focusing on mid-to-late layers for model steering. The rationale is that early layers primarily shape broad syntactic and structural features, well before semantics, knowledge retrieval, and reasoning processes emerge in the mid layers; also, late layers and especially the unembedding layer directly influence the logits, which bias outputs without meaningfully altering intermediate processing steps.

\subsection{Selecting the Steering Direction} 

Our goal is to choose the vector that can  effectively steer the model toward the desired preference-aligned behavior, i.e., favoring human-based preferences in generating responses, while avoiding disruption of its general capabilities.

First, we consider the average signal of residual stream activations induced by the chosen responses from $\mathcal{D}$. Specifically, given   layer $\ell$, we compute the mean residual stream activation at each selected token position, then we take the average across all such positions. We denote the result as~$\bm{\mu}_{\ell}^{(+)}$.

Let us denote  with  $\mathcal{C} = \{ (i,\ell) \}$ the set of pairs (token-position, layer id) that are selected as a result of  the previous stage of position and layer selection.   
We aim to select the steering direction by finding the preference vector $\mathbf{r}_{i, \ell}$ (with $(i,\ell) \in  \mathcal{C}$) that is most strictly aligned with $\bm{\mu}_{\ell}^{(+)}$, that is, \emph{the vector   $\mathbf{r}_{i, \ell}$ that preserves the direction of $\bm{\mu}_{\ell}^{(+)}$ exactly, while  boosting its magnitude.}

This can simply be accomplished by finding the  vector projection onto $\bm{\mu}_{\ell}^{(+)}$ of a $\mathbf{r}_{i, \ell}$  with the maximum magnitude \textit{and} along the direction of $\bm{\mu}_{\ell}^{(+)}$, not backwards. Therefore, by restricting the search to  $\mathbf{r}_{i,\ell} \cdot \bm{\mu}_{\ell}^{(+)} > 0$, we find: 
\begin{equation}
\begin{aligned}
    \mathbf{r}^* &= \arg\max_{(i,\ell) \in  \mathcal{C}} \left\Vert \bigg( \mathbf{r}_{i,\ell} \cdot \dfrac{\bm{\mu}_{\ell}^{(+)}}{\Vert\bm{\mu}_{\ell}^{(+)}\Vert} \bigg) \dfrac{\bm{\mu}_{\ell}^{(+)}}{\Vert\bm{\mu}_{\ell}^{(+)}\Vert}\right\Vert \\
    &= \arg\max_{(i,\ell) \in  \mathcal{C}}  \mathbf{r}_{i,\ell} \cdot \bm{\mu}_{\ell}^{(+)}.
\end{aligned}
\end{equation}

Once we have best aligned a preference direction with the direction of the mean activations of chosen responses, we 
rescale $\mathbf{r}^*$ so that its norm matches $||\bm{\mu}_{\ell^*}^{(+)}||$, where $\ell^*$ denotes the layer corresponding to one of the selected $\mathbf{r}^*$:
\begin{equation}
\mathbf{\hat{r}^*} = \dfrac{||\bm{\mu}_{\ell^*}^{(+)}||}{||\mathbf{r}^*||} \cdot \mathbf{r}^*. 
\end{equation}
 
The above transformation makes the two vectors comparable on the same scale, which allows us to better control the effect of the multiplicative factor in the activation addition step, as described next.

\subsection{Applying the Steering Direction} 
The selected preference direction $\mathbf{r}^*$  is eventually used to steer the model toward preference-aligned behavior.  
As discussed in Sect. \ref{sec:preliminaries}, the model steering is performed through activation addition, which means that the selected preference direction is added to the residual stream activations of any newly generated response by the model. 
Specifically, given an input token sequence $\mathbf{t}$ and by denoting with $\mathbf{x}_{\ell^*}$ the residual stream activations at any token position and at layer $\ell^*$, the steered residuals are defined as follows: 
\begin{equation}
\label{eq:app}
    \mathbf{x}'_{\ell^*} := \mathbf{x}_{\ell^*}(\mathbf{t}) + \alpha \mathbf{\hat{r}}^*,
\end{equation}
where $\alpha \in \mathbb{R}^+$ is a coefficient that controls the strength of the steering effect.  
It should be emphasized that   $\alpha$ is strictly positive by design, since the model steering towards preferred behaviors is conceived to be accomplished through activation addition, rather than ablation---\textit{we aim to encourage desired outcomes, not remove undesired outcomes. } 
Moreover, our empirical evidence indicates that $\alpha$ requires tuning within a narrow range for practical application.

\begin{figure}[t!]
    \centering
    \includegraphics[width=\linewidth]{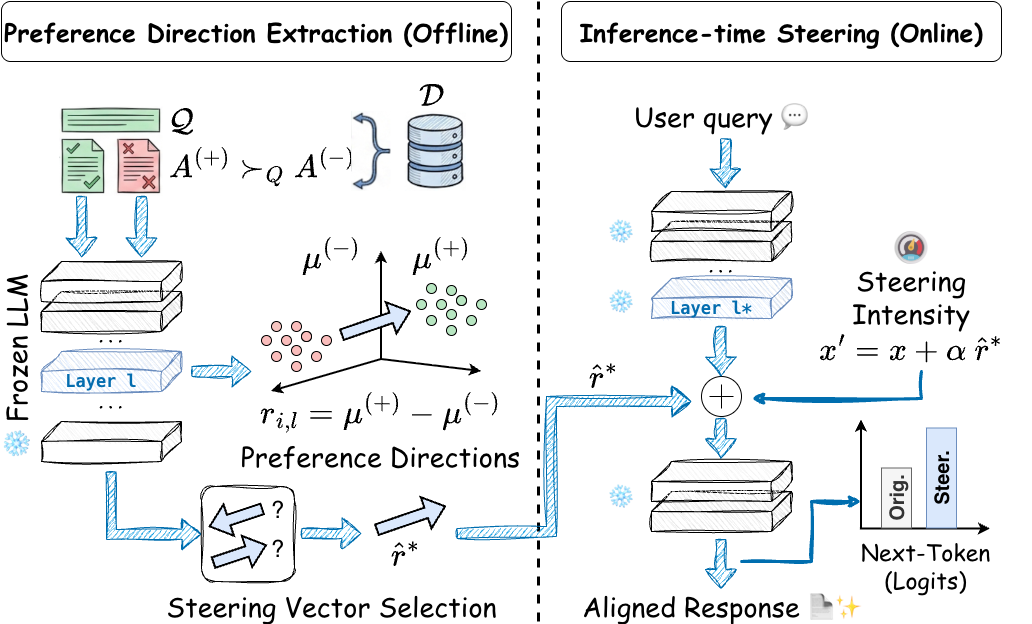}
    \caption{Overview of the main flows  and data modules in \mytool.}
    \label{fig:workflow}
\end{figure}

\vspace{1.5mm}
\noindent 
\textbf{Overview. }
An illustrative summary of the methodological steps of \mytool is shown in  Figure \ref{fig:workflow}. 

\section{Experimental Setup}
\label{sec:setup}
\paragraph{Preference datasets.}
To demonstrate our proposed \mytool approach, we chose to focus on two particularly  informative testbeds for alignment, namely \textbf{mathematical reasoning} and \textbf{coding}. 
Compared to broad, general-purpose tasks like commonsense reasoning, math reasoning and coding demand precise logical reasoning and adherence to rules---small alignment errors can directly break correctness. In summary, \textit{our choice for math reasoning and coding is motivated by an intrinsic objectivity in the assessment of the desired outcomes}, as the  models' responses  can usually be evaluated unambiguously as correct or incorrect. 
Within this view, we used   the \href{https://hf.co/datasets/argilla/distilabel-math-preference-dpo}{argilla/distilabel-math-preference-dpo} dataset, which provides chosen/rejected pairs grounded in mathematical correctness, and \href{https://hf.co/datasets/inclusionAI/Ling-Coder-DPO}{inclusionAI/Ling-Coder-DPO},  whose preferences aim at improving correctness in coding generation.

From each dataset, we randomly sampled 100 triplets to construct the collection $\mathcal{D}$, which we use to compute residual stream activations of chosen and rejected responses.  
Note that the decision to sample a relatively small number of instances is deliberate, as this is sufficient to capture the difference-in-means signal, in line with findings from related work~\cite{Arditi24refusal,Wang25surgical}.
In addition, unlike prior work using residual vectors for refusal ablation~\cite{Arditi24refusal,Wang25surgical} or personality steering~\cite{chen2025persona}, 
our approach does not rely on external cues for sample selection, such as evaluation scores, targeted refusal tokens, or LLMs-as-judges.

\paragraph{Evaluation goals and benchmarks.}
We define the following evaluation goals:\\
$\bullet$ \textbf{(E1)} -- Assessing the performance of \mytool-aligned models against baseline models (i.e., not steered)   using two well-established benchmarks for the target tasks: \textsf{GSM8K}~\cite{gsm8k} for mathematical reasoning and \textsf{HumanEval}~\cite{humaneval} for code generation. \\
$\bullet$ \textbf{(E2)} -- Assessing the  performance  of \mytool-aligned models on tasks outside the target domains, using five widely adopted benchmarks: \textsf{ARC-Challenge}~\cite{arcbenchmark} for scientific and commonsense reasoning, \textsf{HellaSwag}~\cite{hellaswag} for physical and social commonsense inference, \textsf{MMLU}~\cite{Hendrycks21mmlu} for broad knowledge and multi-task understanding, \textsf{TruthfulQA}~\cite{lin-etal-2022-truthfulqa} for factuality and truthfulness, and \textsf{WinoGrande}~\cite{winogrande} for coreference resolution and pronoun disambiguation. By treating these benchmarks as \textit{guardrails}, this evaluation aims to measure whether, and to what extent, \mytool-aligned models for math or coding steering are able to preserve general-purpose capabilities of baseline models.\\
$\bullet$ \textbf{(E3)} -- Comparing \mytool-aligned models with DPO-aligned models and SimPO-aligned models on the target tasks, and evaluating their performance on the benchmarks as well as their efficiency.  To avoid disadvantaging DPO and SimPO, which are known to be more data-intensive, we do not restrict them to the same 100 preference pairs used by \mytool. In addition to these 100 pairs, we also experiment with 1,000 and 10,000 preference pairs---sampled incrementally and using the same seeds as \mytool---for these methods, ensuring a fairer and more informative comparison across alignment approaches. 
\\
$\bullet$ \textbf{(E4)} -- Assessing the parameter sensitivity of the steering coefficient and its impact on \mytool in the target tasks.

To ensure reliability and reproducibility in the evaluation of models, we used the well-established \textsf{Language Model Evaluation Harness} framework~\cite{eval-harness} via its \textsf{TinyBenchmarks} tasks~\cite{tinybenchmarks},  which are a curated selection of samples from the aforementioned benchmarks that ensure the same evaluation robustness as the full one, at a reduced temporal cost. 
Note that~\cite{tinybenchmarks} quantify the average estimation error to be up to 2\%. Accordingly, any performance gain consistently exceeding this threshold is regarded as a significant model-improvement, rather than biased by sampling variance.
We will report performance results that     correspond to \textit{exact\_match} for \textsf{GSM8K},  
 \textit{pass@1} for \textsf{HumanEval}, and \textit{accuracy} for the remaining benchmarks.

\paragraph{Models.}
We experimented with   LLMs differing in family, post-training strategy, and parameter size, as summarized in   Table~\ref{tab:models}. 
These include Llama3~\cite{Llama3} in its 1B, 3B, and 8B variants,  Mistral 7B~\cite{mistral}, and OLMo2 7B~\cite{olmo2}.
This variety enables us to assess to some extent the impact of   model architectures, post-training approaches, and sizes on the steering behavior induced by \mytool. 
As mentioned in Sect. \ref{subsec:extraction}, we inspected mid-to-late layers in each of the models. 
Specifically, we selected a range   $[0.3L..0.9L]$,  where $L$ is the number of layers as reported in Table~\ref{tab:models}.

\begin{table}[t!]
    \centering
    \scalebox{0.8}{
    \begin{tabular}{l|l|c|c}
    \toprule
        \textbf{Model} & 
        \textbf{Param.} & \textbf{Post-training Strategy} & \textbf{No. Layers}  \\
    \midrule
        Llama 1B & 
        1B & SFT + RLHF & 16\\
        Llama 3B & 
        3B & SFT + RLHF & 28\\
        Llama 8B &  
        8B & SFT + RLHF & 32 \\
        Mistral &   
        7B & SFT & 32 \\
        OLMo & 
        7B & SFT + DPO + RLVR & 32 \\
    \bottomrule
    \end{tabular}
    }
    \caption{Summary of used LLMs. Post-training strategies are abbreviated as   SFT: Supervised Fine-Tuning, RLHF: Reinforcement Learning From Human Feedback, RLVR: Reinforcement Learning with Verifiable Reward, DPO: Direct Preference Optimization.}
    \label{tab:models}
\end{table}

\section{Results}

\begin{table}[t!]
    \setlength{\tabcolsep}{1.5pt}
    \renewcommand{\arraystretch}{1.2}
    \centering
    \scalebox{0.7}{
    \begin{tabular}{ll|cc|ccccc}
    \toprule
    & \textbf{Tasks} & \multicolumn{2}{c|}{\textbf{Targets}  $\uparrow$} & \multicolumn{5}{c}{\textbf{Guardrails}  $\uparrow$} \\
    \midrule
    \textbf{\ \ \ \ } & \textbf{Variant} & \textsf{GSM8K}  & \textsf{HumanE.} & \textsf{ARC-C}  & \textsf{HellaS.}   & \textsf{MMLU}  & \textsf{TruthQA}  & \textsf{WinoG.}  \\
    \midrule

    \multirow{5}{*}{\rotatebox{90}{\textbf{Llama 1B}}}
      & Baseline & 0.34 & 0.38 & 0.44 & 0.55 & 0.43 & 0.43 & 0.57 \\
      \cline{2-9}
      
      & \multirow{2}{*}{$\mytool_{Math}$} & \textbf{0.41} & 0.41 & 0.42 & 0.54 & 0.44 & 0.42 & 0.58 \\
      & & {\small(\textbf{+20.10\%})} & {\small(+7.89\%)} & {\small(-3.68\%)} & {\small (-1.03\%)} & {\small (+2.39\%)} & {\small(-2.89\%)} & {\small (+0.90\%) } \\ \cline{2-9}

      & \multirow{2}{*}{$\mytool_{Code}$} & 0.31 & \textbf{0.48} & 0.41 & 0.56 & 0.45 & 0.44 & 0.56 \\
      & & {\small (-9.90\%)} & {\small (\textbf{+26.32\%})} & {\small (-6.70\%)} & {\small (+1.95\%)} & {\small (+4.32\%)} & {\small (+3.64\%)} & {\small(-1.47\%) } \\
    \midrule

    \multirow{5}{*}{\rotatebox{90}{\textbf{Llama 3B}}} 
       & Baseline & 0.62 & 0.60 & 0.57 & 0.78 & 0.63 & 0.48 & 0.61 \\ \cline{2-9}
      
      & \multirow{2}{*}{$\mytool_{Math}$} & \textbf{0.70} & 0.56 & 0.54 & 0.73 & 0.62 & 0.45 & 0.62 \\
      & & {\small (\textbf{+13.53\%})} & {\small (-6.67\%)} & {\small (-5.54\%)} & {\small (-5.60\%)} & {\small (-0.85\%)} & {\small(-5.61\%) } & {\small (+1.61\%)} \\ \cline{2-9}

      & \multirow{2}{*}{$\mytool_{Code}$} &  0.57 & \textbf{0.67} & 0.55 & 0.74 & 0.61 & 0.48 & 0.64 \\
      & & {\small (-7.72\%)} & {\small (\textbf{+11.67\%})} & {\small (-3.66\%)} & {\small (-5.19\%)} & {\small(-2.89\%) } & {\small (-0.39\%)} & {\small (+4.52\%)} \\
    \midrule

    \multirow{5}{*}{\rotatebox{90}{\textbf{Llama 8B}}}
       & Baseline & 0.72 & 0.78 & 0.65 & 0.81 & 0.63 & 0.54 & 0.75 \\
      \cline{2-9}
      
      & \multirow{2}{*}{$\mytool_{Math}$} & \textbf{0.82} & 0.77 & 0.63 & 0.81 & 0.63 & 0.55 & 0.73 \\
      & & {\small (\textbf{+13.09\%})} & {\small (-1.28\%)} & {\small (-3.32\%)} & {\small (+0.32\%)} & {\small (-0.12\%)} & {\small (+1.62\%)} & {\small (-1.71\%)} \\ \cline{2-9}

      & \multirow{2}{*}{$\mytool_{Code}$} &  0.76 & \textbf{0.80} & 0.65 & 0.81 & 0.63 & 0.54 & 0.75 \\
      & & {\small (+4.71\%)} & {\small (\textbf{+2.56\%})} & {\small (-0.52\%)} & {\small (+0.00\%)} & {\small (-0.58\%)} & {\small (+0.18\%)} & {\small (+0.50\%)} \\
    \midrule

    \multirow{5}{*}{\rotatebox{90}{\textbf{Mistral}}} 
       & Baseline & 0.45 & 0.15 & 0.64 & 0.84 & 0.64 & 0.61 & 0.76 \\
      \cline{2-9}
      
      & \multirow{2}{*}{$\mytool_{Math}$} & \textbf{0.53} & 0.15 & 0.65 & 0.84 & 0.64 & 0.61 & 0.76 \\
      & & {\small (\textbf{+18.42\%})} & {\small (+0.00\%)} & {\small (+1.64\%)} & {\small (-0.05\%)} & {\small (-0.19\%)} & {\small (-0.09\%)} & {\small (+0.00\%)} \\ \cline{2-9}

      & \multirow{2}{*}{$\mytool_{Code}$} &  0.47 & \textbf{0.23} & 0.65 & 0.84 & 0.64 & 0.61 & 0.75 \\
      & & {\small (+3.84\%)} & {\small (\textbf{+53.33\%})} & {\small (+0.95\%)} & {\small (+0.02\%)} & {\small (-0.10\%)} & {\small (-1.46\%)} & {\small (-1.31\%)} \\
    \midrule

    \multirow{5}{*}{\rotatebox{90}{\textbf{OLMo}}} 
       & Baseline & 0.75 & 0.52 & 0.66 & 0.83 & 0.62 & 0.56 & 0.76 \\
      \cline{2-9}
      
      & \multirow{2}{*}{$\mytool_{Math}$} & \textbf{0.77} & 0.53 & 0.65 & 0.83 & 0.61 & 0.54 & 0.77 \\
      & & {\small (\textbf{+3.71\%})} & {\small (+1.92\%)} & {\small (-1.19\%)} & {\small (-0.18\%)} & {\small (-1.51\%)} & {\small (-2.43\%)} & {\small (+0.72\%)} \\ \cline{2-9}

      & \multirow{2}{*}{$\mytool_{Code}$} &  0.78 & \textbf{0.60} & 0.66 & 0.83 & 0.62 & 0.55 & 0.75 \\
      & & {\small (+4.88\%)} & {\small (\textbf{+15.38\%})} & {\small (+0.26\%)} & {\small (+0.17\%)} & {\small (+0.00\%)} & {\small (-1.71\%)} & {\small (-1.28\%)}  \\
    \bottomrule    
    \end{tabular}
    }
    \caption{Benchmark performance results: Baseline   vs. \mytool-aligned models. Bold values correspond to \mytool-aligned model performances on the target tasks.
    Values in parenthesis indicate percentage change of a \mytool-aligned model w.r.t. the baseline performance   on the same task. }
    \label{tab:results}
    \vspace{-2mm}
\end{table}

\paragraph{Performance of \mytool on target tasks (E1).} 
Table~\ref{tab:results} shows the benchmark results, where  \mytool-aligned models correspond to the  best settings (cf. \textsl{Suppl.~Mat. B} for details).  
Looking at the two `Targets' columns in the table,  
$\mytool_{Math}$   boosts 
the math-related task (\textsf{GSM8K}) across all models, with an average improvement over the baseline around +14\%  (from +3.7\% with OLMo up to +20.1\% with Llama 1B). 
Analogously, $\mytool_{Code}$  consistently improves  
the code-related task (\textsf{HumanEval}) across all models, with an average improvement over the baseline around     +22\%  (from +2.6\% with Llama 8B up to +53.3\% with Mistral). 

\paragraph{Performance of \mytool on guardrail tasks (E2).}
Valuable insights also arise from performance on the guardrail benchmarks (right-most five columns in Table~\ref{tab:results}).  
  \mytool-aligned 
Llama 1B, Llama 8B,  and OLMo are fairly stable, with average percentage changes on guardrail benchmarks around -1\% or less:  for $\mytool_{Math}$ resp. $\mytool_{Code}$,  
\text{-0.86}\% resp. +0.35\% with Llama 1B, -0.64\% resp. \text{-0.08}\% with Llama 8B, and   
-0.73\% resp. -0.51\% with OLMo. 
$\mytool_{Math}$ with Mistral even slightly improves on average over the guardrails (+0.26\%), while keeping average guardrail degradation around -0.38\% when using $\mytool_{Code}$.  
By contrast, Llama 3B is the only   to suffer the most guardrail degradation, up to -3.19\% with $\mytool_{Math}$.

Regarding the model types, Mistral has the largest boost by $\mytool_{Code}$ (+53.4\%) and second-largest by $\mytool_{Math}$ (+18.4\%). 
Using Llama models has shown a scaling effect, since smaller models lead to relatively larger gains but also stronger impact on guardrail tasks.   \mytool-aligned OLMo models have the least improvement (+3.7\%) over the baseline on the math task, but +15.4\% on coding task. 

Remarkably, \mytool alignment of the largest models, i.e., OLMo, Mistral and Llama 8B, on one target task reveals little degradation  or, more often, improvement over the other target task, suggesting that steering in particular on a math task can have a beneficial effect on a coding task as well.

\begin{figure}[t!]
    \centering
    \begin{tabular}{c}      
    \includegraphics[width=0.7\linewidth]{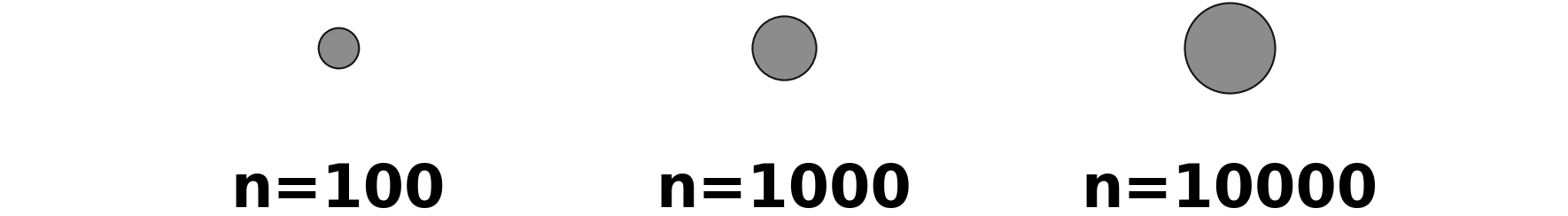} \\
    \includegraphics[width=0.95\linewidth]{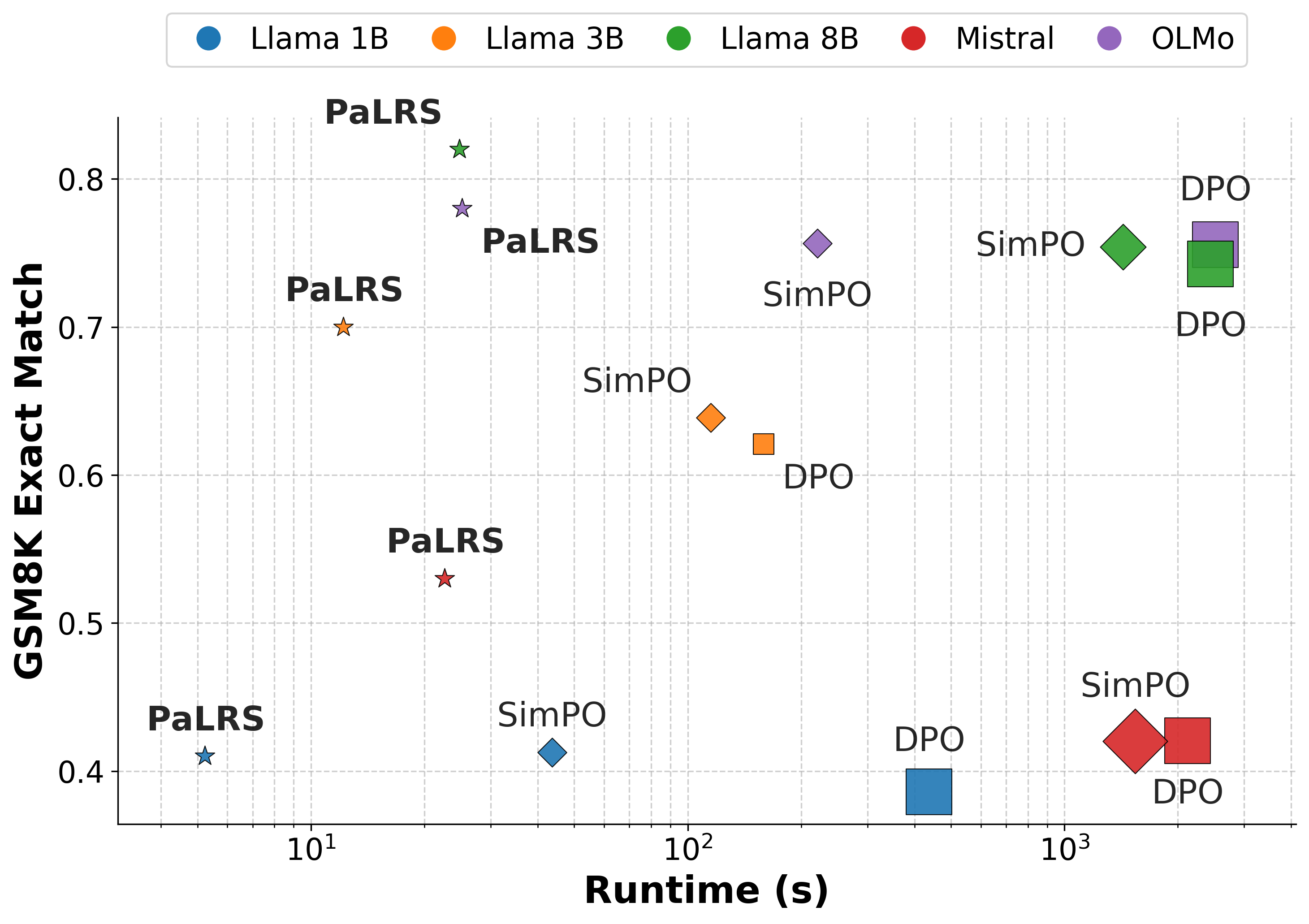} 
         \\
\includegraphics[width=0.95\linewidth]{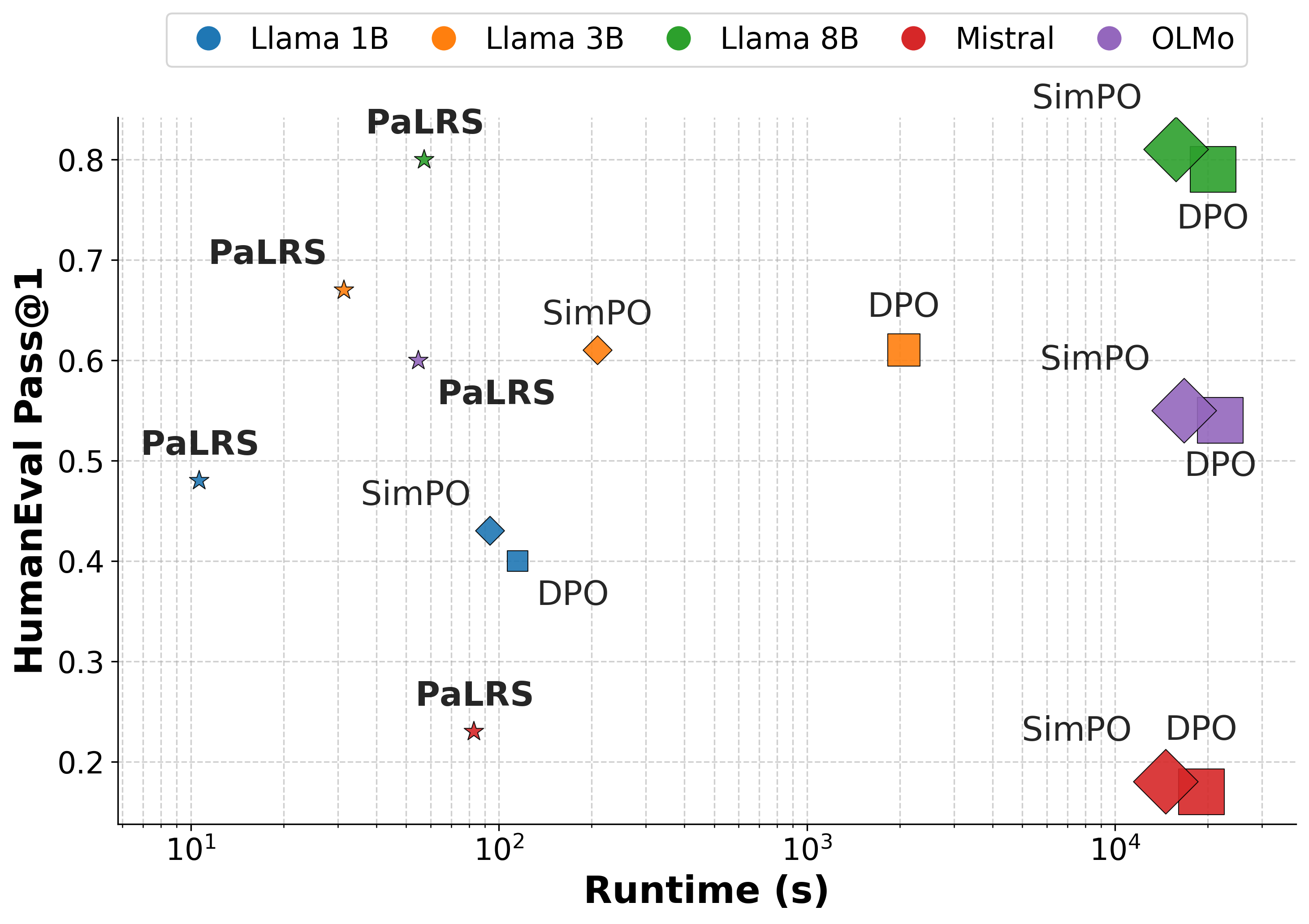}
\vspace{-3mm}
\end{tabular}
 \caption{Best-setting benchmark performance and efficiency comparison of \mytool-aligned models vs. DPO-aligned and SimPO-aligned models, for different numbers ($n$) of preference pairs: \textsf{GSM8K} (top) and \textsf{HumanEval} (bottom). Colors denote models, while stars, resp. circles and squares,  denote \mytool-alignment, resp. other  alignments. Marker size increases with $n$.}
    \label{fig:comparison-dpo-efficiency}
\end{figure}

\begin{figure*}[t!]
    \centering
\includegraphics[width=0.98\linewidth]{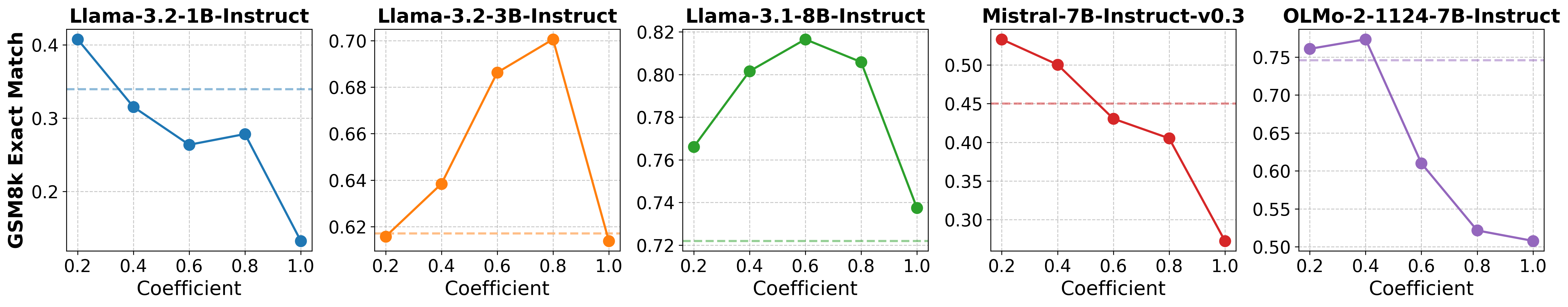} \\
    \includegraphics[width=0.98\linewidth]{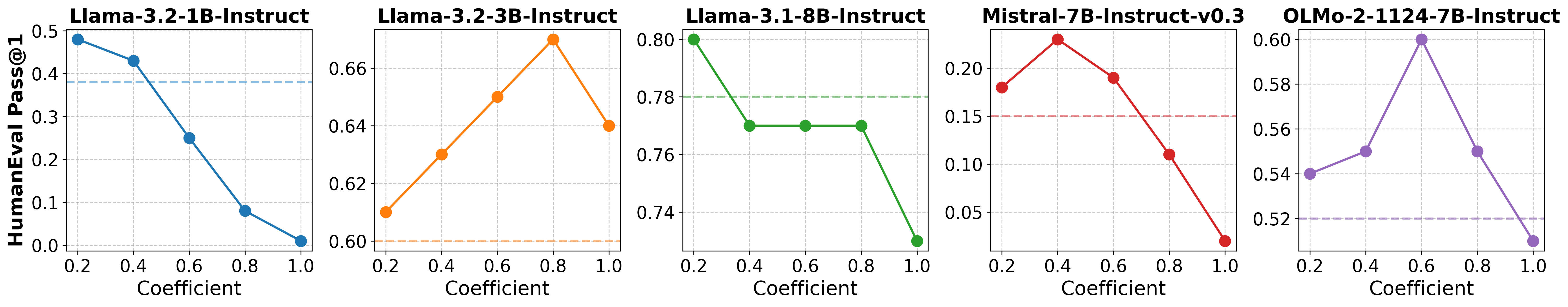} \\
    \caption{Performance of \mytool-aligned models on the   target tasks \textsf{GSM8K} (top) and  \textsf{HumanEval} (bottom)  by varying the steering coefficient. Dashed horizontal line marks the baseline performance.}
    \label{fig:alpha-sensitivity}
\end{figure*}
 
\paragraph{Comparison of \mytool with DPO and SimPO (E3).} 
Figure \ref{fig:comparison-dpo-efficiency} compares the best-setting  \mytool-aligned models with DPO-  and SimPO-aligned models, in terms of effectiveness and efficiency on the \textsf{GSM8K} and \textsf{HumanEval} benchmarks, by varying model size and number $n$ of preference pairs for the alignment.  DPO and SimPO based values correspond to the best benchmark scores across various alignment-data sizes $n$---details are reported in \textsl{Suppl.~Mat. C}---while for \mytool  $n$ is kept   fixed to 100 (Sect.~\ref{sec:setup}).

On \textsf{GSM8K}, $\mytool_{Math}$-aligned models consistently surpass   the corresponding DPO- and SimPO-aligned models, across  model types (the sole exception is Llama 1B, for which   $\mytool_{Math}$ ties with SimPO) with performance gains:  +13\% (Llama 3B), +10\% (Llama 8B), +29\% (Mistral), +2\% (OLMo). 
 These results couple with the  outstanding  time efficiency of  $\mytool_{Math}$-aligned models, which  are 1 or 2  orders of magnitude faster than  DPO-aligned models:   for example, using Llama 8B, learning the $\mytool_{Math}$-aligned model takes about 25s vs. DPO-aligned one’s  2444s, i.e., $\sim 98$x faster.   
Even more remarkably, the efficiency advantage taken by \mytool  also holds in comparison with SimPO-aligned models, and is particularly evident  for larger models (e.g., $\sim 57$x faster for Llama 8B.)

Analogous remarks can be drawn from the comparison on     \textsf{HumanEval}, where performance gains of \mytool vs. SimPO---which in this benchmark behaves consistently better or on par with DPO---are  +11.6\% (Llama 1B), +9.8\% (Llama 3B), +0.0\% (Llama 8B), +27.8\% (Mistral), +9.1\% (OLMo).   
 Compared to \textsf{GSM8K},  $\mytool_{Code}$-aligned models' efficiency is even more marked, up to 3 orders of magnitude faster than both      DPO-aligned  and SimPO-aligned models, as shown in the figure for   Llama 8B, OLMo, and Mistral.

\paragraph{Impact of the steering coefficient (E4).}
We investigate the sensitivity of the steering coefficient $\alpha$ (cf. Eq. (\ref{eq:app})) and its effect on the performance of \mytool-aligned models on   the target tasks. 
Figure \ref{fig:alpha-sensitivity} provides insights which can be summarized as follows.  We notice that the coefficient sensitivity is model- and task-dependent. While in general performance trends  drop consistently as the coefficient increases, there is always a regime of the coefficient where the \mytool-aligned models outperform the relative baselines. 
Low to moderate   coefficients (up to  0.8) often yield the best trade-offs, especially for Llama 3B, Llama 8B (\textsf{GSM8K}) and OLMo (\textsf{HumanEval}). 
Oversteering is most visible at coefficient 1.0 for all models. More specifically, for Llama 3B and OLMo  (\textsf{HumanEval}), the beneficial range is around 0.6–0.8 before oversteering occurs; for Mistral  and Llama 1B, oversteering happens much earlier (as low as 0.4–0.6), while Llama 8B tolerates higher coefficients better but still shows oversteering when pushed too far.  
To sum up, steering is actually    beneficial, although, to avoid oversteering and performance degradation,  $\alpha$ needs   tuning, e.g., 
 by leveraging a small held-out split of the preference dataset using the same metrics as the target benchmark.

\section{Conclusions}
In this work, we presented \mytool, a training-free method for preference alignment that leverages residual stream activations to steer LLM behavior directly at inference time. By distilling steering vectors from as few as a hundred preference pairs, \mytool aligns models with desired behaviors without any parameter updates, costly optimization, or the need to maintain multiple fine-tuned checkpoints. Crucially, our results show that \mytool-aligned models are a safer and consistently superior alternative to their DPO- and SimPO-aligned counterparts: never underperforming, often achieving substantial gains, and doing so with orders-of-magnitude less computation. Our findings render residual-based steering as a powerful paradigm for preference alignment, aiming to make it simpler, more effective yet scalable, and broadly accessible compared to traditional post-training approaches.

\paragraph{Limitations.}
Our results have highlighted the promise of residual-based preference alignment, through the evaluation of  \mytool on a relatively representative sample of model families and post-training modalities. Despite this,  generalizability to very-large-scale   models needs to be evaluated. 
Our current method for discovering effective data subsets and steering coefficients relies on heuristic grid search: developing principled, theoretically grounded selection strategies remains a key direction for smoother real-world deployment of our results. 
Yet,  while our results demonstrate that residual interventions effectively steer model behavior and provide empirical support for our hypothesis, more in-depth explainability of how preference information is encoded and disentangled across layers is needed.

\vspace{-1mm}
\section*{Ethical Statement}
While  highlighting the potential for beneficial applications, we acknowledge that our work   might also be misused to steer models toward harmful or malicious behaviors. We strongly discourage any such misuse and do not assume responsibility for applications beyond the scope of this research.

\clearpage

\appendix
\twocolumn[{%
\centering
\begin{center}

\textbf{\LARGE Technical Appendix}

\ \\

\ \\
	
\end{center}
}]

\section{Running Environment}
Our experiments were carried out on a machine with 8x 24GB-RAM NVIDIA A30 GPU, 764GB RAM, Double Intel Xeon Gold 6248R with  96 cores,   Ubuntu Linux 20.04.6 LTS.

\section{Details on the Model Configurations Used for the Main Results}
\label{app:best-config}
Table~\ref{tab:configs} reports the best model configurations used to obtain $\mathbf{\hat{r}^*}$, and the corresponding steering coefficient $\alpha$, used for the results shown throughout the main paper.

\begin{table}[h]
    \centering
    \scalebox{0.87}{
    \begin{tabular}{l|cccc|cccc}
    \toprule
    & \multicolumn{4}{c|}{\textbf{Math}} & \multicolumn{4}{c}{\textbf{Code}}\\
    \midrule
    \textbf{Model} & seed & \textbf{$i$} & \textbf{$\ell^* / L$} & \textbf{$\alpha$} & seed & \textbf{$i$} & \textbf{$\ell^* / L$} & \textbf{$\alpha$}\\
    \midrule
        Llama 1B & 870& -2 & 14/16 & 0.2          & 343 & -5 & 14/16 & 0.2 \\
        Llama 3B & 921& -5 & 23/28 & 0.8          & 689& -1 & 24/28 & 0.8 \\
        Llama 8B & 245& -4 & 28/32 & 0.6          & 94& -4 & 26/32 & 0.2 \\
        Mistral  & 790& -2 & 24/32 & 0.2          & 237& -1 & 28/32 & 0.4 \\
        OLMo     & 311& -3 & 23/32 & 0.4          & 447& -3 & 26/32 & 0.6 \\
    \bottomrule
    \end{tabular}
    }
    \caption{(Grid-search based) Best seed, token position ($i$), layer ($\ell^*/L$) and steering coefficient ($\alpha$) used throughout our experimental evaluation of \mytool-aligned models.}
    \label{tab:configs}
\end{table}

\section{Hyperparameters for DPO and SimPO Alignment}
\label{app:sec:dposimpoparams}
Table~\ref{tab:dpo_hyperparams}, resp. Table~\ref{tab:simpo_hyperparams} reports the settings we used to align our considered models with DPO, resp. SimPO, for a comparison with \mytool.
Both methods are configured following established best practices from NVIDIA,\footnote{\url{https://github.com/NVIDIA/NeMo-Aligner}} the TRL library,\footnote{\url{https://github.com/huggingface/trl}} and following the original implementations.

As a key challenge in preference optimization is adapting training dynamics to dataset size. We employ a principled scaling strategy. For small datasets ($N=100$), we perform multiple epochs (3) with higher learning rates and stronger regularization to extract maximum signal while preventing overfitting. For medium ones ($N=1000$), we adopt a more balanced setting with 2 epochs and moderate effective batch sizes (128). For large datasets ($N=10000$), we perform a single-epoch training with large batch sizes (256) and conservative learning rates to prevent catastrophic drift from the base model.

For DPO, we keep the default paper $\beta = 0.1$~\cite{dpo} for small and medium datasets. For larger datasets, we increase $\beta$ to 0.2 following NVIDIA's guidance that higher values help prevent reward hacking and maintain output quality when training on more diverse preference pairs. 

For SimPO, we set $\beta = 2.0$ and $\gamma = 1.0$, maintaining the recommended ratio $\gamma/\beta \approx 0.5$ which was shown to achieve optimal performance across benchmarks.\footnote{\url{https://github.com/princeton-nlp/SimPO}}

Tables~\ref{tab:dpo_hyperparams} and~\ref{tab:simpo_hyperparams} detail the complete hyperparameter configurations for DPO and SimPO, respectively.

\begin{table}[h!]
\centering
\scalebox{0.85}{
\begin{tabular}{lccc}
\toprule
\textbf{Hyperparameter} & \textbf{N=100} & \textbf{N=1,000} & \textbf{N=10,000} \\
\midrule
Batch size (per device)        & 2 & 2 & 2 \\
Gradient accumulation steps    & 2 & 8 & 16 \\
Effective batch size           & 32 & 128 & 256 \\
Learning rate                  & $9\text{e-}7$ & $5\text{e-}7$ & $5\text{e-}7$ \\
Scheduler                      & Cosine & Cosine & Cosine \\
Warmup steps                   & 5 & 10 & 20 \\
Optimizer                      & AdamW  & AdamW  & AdamW  \\
Training epochs                & 3 & 2 & 1 \\
$\beta$ coefficient            & 0.1 & 0.1 & 0.2 \\
Weight decay                   & 0.1 & 0.05 & 0.01 \\
Max gradient norm              & 1.0 & 1.0 & 1.0 \\
Precision                      & bfloat16 & bfloat16 & bfloat16 \\
\bottomrule
\end{tabular}
}
\caption{Training hyperparameters used for DPO at different dataset scales.}
\label{tab:dpo_hyperparams}
\end{table}

\begin{table}[h!]
\centering
\scalebox{0.85}{
\begin{tabular}{lccc}
\toprule
\textbf{Hyperparameter} & \textbf{N=100} & \textbf{N=1,000} & \textbf{N=10,000} \\
\midrule
Batch size (per device)        & 2 & 2 & 2 \\
Gradient accumulation steps    & 2 & 8 & 16 \\
Effective batch size           & 32 & 128 & 256 \\
Learning rate                  & $1\text{e-}5$ & $5\text{e-}6$ & $1\text{e-}6$ \\
Scheduler                      & Cosine & Cosine & Cosine \\
Warmup ratio                   & 0.03 & 0.03 & 0.03 \\
Optimizer                      & AdamW  & AdamW  & AdamW  \\
Training epochs                & 3 & 2 & 1 \\
$\beta$ coefficient            & 2.0 & 2.0 & 2.0 \\
$\gamma$ (SimPO margin)        & 1.0 & 1.0 & 1.0 \\
Weight decay                   & 0.05 & 0.02 & 0.01 \\
Max gradient norm              & 1.0 & 1.0 & 1.0 \\
Precision                      & bfloat16 & bfloat16 & bfloat16 \\
\bottomrule
\end{tabular}
}
\caption{Training hyperparameters used for SimPO at different dataset scales.}
\label{tab:simpo_hyperparams}
\end{table}

\section{Qualitative Examples of \mytool's Steering Effect}
Figures~\ref{fig:math-example1}-\ref{fig:code-example2} provide qualitative insights into the effect of model steering toward preferred behaviors obtained via \mytool.

\begin{figure}[ht!]
    \centering
    \begin{tcolorbox}[colback=white!5!white,colframe=gray!75!black, width=\linewidth]

        \textbf{Question:} \textit{Mia and Emma are currently 16 years apart in age. If Mia, who is younger than Emma, is 40 years old, what's the average of their ages?}

        \medskip
        \textbf{Expected Answer:} \\
        If Mia is 40 years old, Emma is $40 + 16 = \textbf{56}$ years old. \\
        The sum of their ages is $56 + 40 = \textbf{96}$ years. \\
        The average age of the two is $96 / 2 = \textbf{48}$ years.  

        \medskip
        
        \textbf{Baseline:} Mia is 40 years old and Emma is \textcolor{red}{\textbf{16 years old}}, so the average age is $(40 + \textcolor{red}{\textbf{16}}) / 2 = \textcolor{red}{\textbf{56 / 2 = 28}}$ years.

        \medskip
        \textbf{\mytool$_{Math}$:} Mia is 40 years old and Emma is $40 + 16 = \textcolor{teal}{\textbf{56}}$ years old. \\
        The average age of the two of them is $(40 + \textcolor{teal}{\textbf{56}}) / 2 = \textcolor{teal}{\textbf{96 / 2 = 48}}$ years.

    \end{tcolorbox}
    \caption{Example of mathematical improvement via \mytool for Llama 3.2 1B.}
    \label{fig:math-example1}
\end{figure} 

\begin{figure}[ht!]
    \centering
    \begin{tcolorbox}[colback=white!5!white,colframe=gray!75!black, width=\linewidth]

        \textbf{Question:} \textit{James has 30 teeth. His dentist drills 4 of them and caps 7 more teeth than he drills. What percentage of James' teeth does the dentist fix?}
        
        \medskip
        \textbf{Expected Answer:} \\
        First find the total number of teeth the dentist caps: 4 teeth + 7 teeth = \textbf{11 teeth} \\
        Then find the total number of teeth the dentist fixes: 11 teeth + 4 teeth = \textbf{15 teeth} \\
        Then divide that number by the total number of teeth and multiply by 100\% to find the percentage of teeth he fixes: 15 teeth / 30 teeth * 100\% = \textbf{50\%}  

        \medskip
        
        \textbf{Baseline:} James has 30 teeth. The dentist drills 4 teeth. \textcolor{red}{\textbf{So, the number of teeth that the dentist fixes is 30 - 4 = 26.}} \\
        The dentist caps 7 more teeth than he drills. So, the number of teeth that the dentist caps is 4 + 7 = 11. \\
        The number of teeth that the dentist fixes is \textcolor{red}{\textbf{26}}. The number of teeth that the dentist caps is 11. So, the percentage of teeth that the dentist fixes is \textcolor{red}{\textbf{26 / 30 * 100 = 86.67\%}}  
        
        \medskip
        \textbf{\mytool$_{Math}$:} James has 30 teeth. His dentist drills 4 teeth, so he has 30 - 4 = 26 teeth left. \\
        The dentist caps 7 more teeth than he drills, so he caps 4 + 7 = \textcolor{teal}{\textbf{11 teeth}}. \\
        So James has 26 - 11 = \textcolor{teal}{\textbf{15 teeth}} that the dentist fixes. \\
        The percentage of teeth that the dentist fixes is \textcolor{teal}{\textbf{15/30 * 100\% = 50\%}}  
            
    \end{tcolorbox}
    \caption{Example of mathematical improvement via \mytool for Llama 3.2 1B.}
    \label{fig:math-example2}
\end{figure}

\begin{figure}[ht!]
    \centering
    \begin{tcolorbox}[colback=white!5!white,colframe=gray!75!black, width=\linewidth]

    \lstset{
    basicstyle=\ttfamily\small,
    breaklines=true,
    breakatwhitespace=true,
    columns=fullflexible,
    frame=none,
    tabsize=4
}

        \textbf{Question:} 
        \begin{lstlisting}
def count_distinct_characters(string: str) -> int:
    """ Given a string, find out how many distinct characters (regardless of case) does it consist of
    
    count_distinct_characters(`xyzXYZ')
    >>> 3
    count_distinct_characters(`Jerry')
    >>> 4
    """
\end{lstlisting}

        \medskip
        
        \textbf{Baseline:}
        \begin{lstlisting}
def count_distinct_characters(string: str) -> int:
   # Convert the string to lowercase and remove any whitespace
    string = string.lower().replace(" ", "")

    # Create a set from the string and return the length
    # of the set
    return len(set(string))
\end{lstlisting}

        \medskip
        \textbf{\mytool$_{Code}$:} 
        \begin{lstlisting}
def count_distinct_characters(string: str) -> int:
   # Create a set to store the characters
    char_set = set()

    # Iterate through the string and add each character
    # to the set
    for char in string:
        char_set.add(char.lower())

    # Return the size of the set
    return len(char_set)
\end{lstlisting}

    \end{tcolorbox}
    \caption{Example of coding improvement via \mytool for Mistral 7B.}
    \label{fig:code-example1}
\end{figure}

\begin{figure}[ht!]
    \centering
    \begin{tcolorbox}[colback=white!5!white,colframe=gray!75!black, width=\linewidth]

    \lstset{
    basicstyle=\ttfamily\small,
    breaklines=true,
    breakatwhitespace=true,
    columns=fullflexible,
    frame=none,
    tabsize=4
}
        
        \textbf{Question:} 
        \begin{lstlisting}
def string_xor(a: str, b: str) -> str:
    """ Input are two strings a and b consisting 
    only of 1s and 0s.
    Perform binary XOR on these inputs and return 
    result also as a string.
    >>> string_xor('010', '110')
    '100'
    """
\end{lstlisting}

        \medskip
        
        \textbf{Baseline:}
        \begin{lstlisting}
def string_xor(a: str, b: str) -> str:
   result = ""
   for i in range(len(a)):
        if a[i] == b[i]:
            result += "0"
        else:
            result += "1"
    return result
\end{lstlisting}

        \medskip
        \textbf{\mytool$_{Code}$:} 
        \begin{lstlisting}
def string_xor(a: str, b: str) -> str:
   # Initialize result string
    result = ""

    # Iterate through both strings
    for i in range(len(a)):
        # Perform XOR operation
        result += str(int(a[i]) ^ int(b[i]))

    return result
\end{lstlisting}

    \end{tcolorbox}
    \caption{Example of coding improvement via \mytool for Mistral 7B.}
    \label{fig:code-example2}
\end{figure}

\end{document}